\documentclass[11pt]{article}
    
\usepackage[final]{acl} 
\usepackage{graphicx} 
\usepackage{subcaption} 
\usepackage{cuted}
\usepackage{tcolorbox}
\usepackage{dashrule}
\usepackage{amsmath, amssymb} 
\usepackage{booktabs}
\usepackage{multirow}
\usepackage{times}
\usepackage{latexsym}
\usepackage{amsmath}
\usepackage{cuted} 
\usepackage{color}
\usepackage{xspace}

\usepackage{colortbl}
\definecolor{mygray}{gray}{0.8}

\usepackage{times}
\usepackage{enumitem}

\usepackage{hyperref}
\usepackage{xcolor}
\usepackage{fontawesome5}  

\newcommand{\one}{({\em i}\/)\xspace}
\newcommand{\two}{({\em ii}\/)\xspace}
\newcommand{\three}{({\em iii}\/)\xspace}

\newtcolorbox{examplebox}[1]{
  colback=blue!3!white,
  colframe=blue!3!gray,
  fonttitle=\bfseries\sffamily,
  coltitle=white,
  title={#1},
  arc=2pt,
  boxrule=0.8pt,
  left=6pt, right=6pt, top=4pt, bottom=4pt,
  width=\linewidth
}

\newtcolorbox{exampleboxer}[1]{
  colback=blue!3!white,
  colframe=blue!3!black,
  fonttitle=\bfseries\sffamily,
  coltitle=white,
  title={#1},
  arc=2pt,
  boxrule=0.8pt,
  left=6pt, right=6pt, top=4pt, bottom=4pt,
  width=\linewidth
}

\usepackage{inconsolata}

\title{CogniDir: Combating Cognitive Malicious Comments via Adaptive Distributional Learning for Robust Fake News Detection}

\author{
  \textbf{Zhao Tong$^{1,2}$\footnotemark[1], Chunlin Gong $^{3}$\footnotemark[1], Yimeng Gu$^{4}$, Haichao Shi$^{1}$, Qiang Liu$^{5}$} ,\\ \textbf{Shu Wu$^{5}$\footnotemark[2], Xingcheng Xu$^{6}$}, 
  \textbf{Xiao-Yu Zhang$^{1}$\footnotemark[2]}\\
  $^1$Institute of Information Engineering, Chinese Academy of Sciences \\
  $^2$School of Cyber Security, University of Chinese Academy of Sciences \\
  $^3$University of Minnesota,
  $^4$Queen Mary University of London\\ 
  $^5$New Laboratory of Pattern Recognition (NLPR), \\
  State Key Laboratory of Multimodal Artificial Intelligence Systems (MAIS), \\
  Institute of Automation, Chinese Academy of Sciences 
  \\  $^6$Shanghai AI Laboratory\\
  \texttt{tongzhao@iie.ac.cn, gong0226@umn.edu
  }\\
  \texttt{shu.wu@nlpr.ia.ac.cn, zhangxiaoyu@iie.ac.cn}\\
{
\raisebox{-0.10em}{\faGlobe}\,
\href{https://chunlingong.me/projects/cognidir/}{\textbf{Project Page}}
\hspace{0.8em}\textbar\hspace{0.8em}
\raisebox{-0.10em}{\faGithub}\,
\href{https://github.com/cheslyn0712/CogniDir}{\textbf{GitHub Code}}
}\\
}

\begin{document}
\maketitle

\renewcommand{\thefootnote}{\fnsymbol{footnote}}
\footnotetext[1]{These authors contributed to the work equally.}
\footnotetext[2]{To whom correspondence should be addressed.} 
\renewcommand{\thefootnote}{\arabic{footnote}}

\begin{abstract}
The proliferation of Large Language Models (LLMs) has enabled a new class of psychologically grounded malicious comments, shifting fake news attacks from surface-level textual noise to deep cognitive and logical manipulation. This shift severely undermines existing detectors, which conventionally rely on static attack assumptions and fixed training distributions. To bridge this gap, we introduce \texttt{CogniDir}, an adaptive distributional learning framework that reformulates robust detection as a dynamic data mixture optimization problem for social media content safety. Grounded in cognitive psychology, we first formalize mechanism-specific cognitive adversarial paradigms to systematically expose deep-seated detector vulnerabilities. To address the vulnerability heterogeneity, CogniDir derives an information-theoretic score coupling empirical accuracy with probabilistic confidence, which is then mapped to adaptive sampling proportions through a Dirichlet-mean parameterization. This formulation enables smooth, feedback-driven reallocation of training exposure toward the most brittle attack mechanisms. Experimental results on three benchmarks demonstrate that CogniDir yields state-of-the-art robustness, improving F1 scores by up to 17.9\% over competitive baselines under heterogeneous, AI-generated adversarial pressures.

\end{abstract}


\section{Introduction}
The paradigm of fake news detection has been fundamentally challenged by the rapid proliferation of Large Language Models (LLMs)\citep{tong2026rlshield, tong2026cot}. Beyond generating standard fake news, malicious attackers can now deploy automated, LLM-augmented user comments that act as sophisticated adversarial perturbations against existing detectors \citep{chen2024combating, nutsugah2025social}. Unlike conventional text perturbations that rely on surface-level lexicon alterations, these LLM-generated malicious comments systematically exploit cognitive and algorithmic blind spots through strategic logical manipulation and psychological framing \citep{tong2025generate, hu2025llm}. Consequently, they induce severe alignment degradation and classification blindness in state-of-the-art detection systems. Maintaining proactive robustness under such heterogeneous, mechanism-driven adversarial pressures has thus emerged as a critical yet under-explored frontier in digital content security.

\begin{figure}[!t]
  \centering
  \includegraphics[width=\columnwidth]{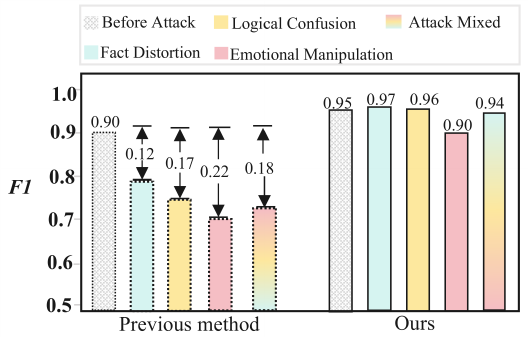}
  \caption{Group-wise F1 comparison on Weibo16\citep{ma2016detecting}. Previous method \citep{shu2019defend} typically suffers a performance drop of more than 18\% under different types of malicious comments. Yet, our model is able to significantly narrow this performance gap, achieving 0.94 in F1 across diverse malicious comments. }
  \label{fig:Teaser_Fig}
  \vspace{-0.2cm}
\end{figure}

To counter these evolving threats, the current detectors leverage crowd intelligence to fortify verification boundaries \citep{guo2022survey, nan2025exploiting}. However, they inevitably suffer from severe \textit{distributional mismatch} under realistic deployment, driven by a continuous causal failure chain: (i) \textbf{Mechanism Homogenization}: Existing frameworks conventionally treat diverse cognitive malicious comments as generic text noise, making them fundamentally incapable of distinguishing fine-grained interference pathways like fact distortion, logical confusion, and emotional manipulation (see Fig.~\ref{fig:Teaser_Fig}). (ii) \textbf{Static Data Assumptions}: Because they fail to decouple these distinct cognitive mechanisms, their subsequent defenses rely entirely on fixed, homogeneous training distributions that cannot track the fluid, evolutionary nature of adaptive LLM strategies. (iii) \textbf{Uniform Training Dynamics}: Crucially, this reliance on invariant data forces the optimization to apply static data mixing across all attack classes, implicitly assuming identical defense difficulty. Consequently, the optimization trajectory rapidly overfits to shallow textual patterns, leaving group-wise vulnerability heterogeneity unaddressed and causing robustness to plateau under novel attack distributions.

To bridge this gap, we propose \textbf{CogniDir}, a framework that reframes robust detection as a \textit{dynamic distributional optimization problem}. Grounded in cognitive psychology~\citep{pennycook2021psychology, park2025confirmation, moradi2021evaluating, yeh2024cocolofa, liu2024conspemollm}, CogniDir addresses two core challenges: First, to overcome the scarcity of mechanism-labeled adversarial data, we design a \textit{cognitive-grounded synthesis pipeline} that leverages LLMs to generate diverse, category-specific perturbations, establishing a scalable foundation for AI safety research. Second, to achieve adaptive robustness, we formalize training exposure allocation as an optimization problem over the probability simplex. By deriving an information-theoretic vulnerability metric that couples accuracy with probabilistic confidence, we map group-wise brittleness to adaptive sampling proportions through a
Dirichlet-mean parameterization, enabling smooth, feedback-driven reallocation of sampling budgets. This transforms adversarial training into a self-correcting curriculum that continuously targets the model’s weakest mechanisms while ensuring cross-lingual transferability.

Our contributions are summarized as follows:
\begin{itemize}
    \item  We introduce a cognitive-grounded adversarial synthesis pipeline to generate mechanism-labeled malicious comments, which improves training data diversity and fidelity.
    \vspace{-0.5em}
    \item  We propose CogniDir, a distributional optimization framework in which an information-theoretic vulnerability metric adaptively maps group-wise vulnerability for feedback-driven allocation.
    \vspace{-0.5em}
    \item We demonstrate that CogniDir achieves state-of-the-art robustness across three cross-lingual benchmarks, narrowing the vulnerability gap under heterogeneous attacks while maintaining high baseline accuracy.
\end{itemize}

\section{Related Work}
\subsection{Comment-based Fake News Detection}
News comments provide crowd intelligence that complements news content for fake news detection \cite{shu2019defend,yang2023rumor,nan2024let}. Early methods co-model news and comments via attention mechanisms to jointly capture check-worthy content \cite{shu2019defend}, sequential features \cite{yang2023rumor}, or crucial comment-context interactions \cite{wang2023comment}. However, diverse comments are not always available. To address this, recent studies explore data augmentation: \cite{nan2024let} use LLMs to generate alternative comments, \cite{wan2024dell} simulate user-news interactions with generated reactions, and \cite{yanagi2020fake} augment sets by generating new comments. Meanwhile, other works deepen comment modeling, utilizing comments to guide text–image cross-modal attention \cite{yi2025user} or modeling interactions between news emotion and comment semantics \cite{tan2026emotion}.

Despite these advances, the vulnerability of comment-based detectors to malicious comments remains understudied. In this work, we systematically evaluate and attack detectors through three categories of comment perturbations, and propose a novel framework to defend against such attacks.


\subsection{Attack on Fake News Detection}

Despite advances in fake news detection, attacking these detectors remains underexplored \cite{le2020malcom,zhu2024generalblackboxadversarialattack,WWW2025_attack}. Existing attacks primarily target news content, propagation structures, or comments. Content- and structure-based attacks modify writing styles \cite{KDD2024_sheepDog}, apply meaning-preserving transformations \cite{przybyla2024attacking}, or perturb propagation networks \cite{zhu2024generalblackboxadversarialattack} and user graphs \cite{wang2024bots}. In contrast, comment-based attacks manipulate user responses while keeping the news unchanged. Specifically, attackers generate malicious comments \cite{le2020malcom} or transfer real comments from other articles \cite{koren2025evaluating}. Consequently, adversarial training has been explored to improve robustness \cite{WWW2025_attack,sun2025unifying}.

Content and comment attacks correspond to distinct attack surfaces and capabilities: modifying publisher-controlled text versus injecting comments in open environments. Focusing on the latter, our work investigates heterogeneous cognitive mechanisms in malicious comments and enhances comment-aware detectors via vulnerability-guided adaptive adversarial training.

\section{Problem Definition}
Comment-based fake news detection predicts the authenticity of a given news based on the news content and its associated comments. Formally, let the news dataset be
$D=\{(x_i,\mathcal{C}_i,y_i)\}_{i=1}^{N}$,
where $x_i$ denotes the content of the $i$-th news article, $\mathcal{C}_i=\{c_i^{j}\}_{j=1}^{M}$ denotes its associated original comment set, and $y_i\in\{0,1\}$ is the news authenticity label ($1$ for fake and $0$ for real).
Given $(x_i,\mathcal{C}_i)$, the goal is to learn a detector $f$ that predicts $y_i$.

During the attack, the news content $x_i$ remains unchanged, while malicious comments $\mathcal{C}_i^{adv}$ are injected into the original comment set, forming
$\mathcal{C}_i'=\mathcal{C}_i\cup\mathcal{C}_i^{adv}$.
For an originally correct prediction $f(x_i,\mathcal{C}_i)=y_i$, an attack succeeds when
$f(x_i,\mathcal{C}_i')\neq y_i$.
We consider untargeted attacks that include both real-to-fake and fake-to-real prediction flips(see Appendix~\ref{app:directional}).

\begin{figure*}[!ht]
    \centering
    \includegraphics[width=\textwidth,height=0.3\textheight]{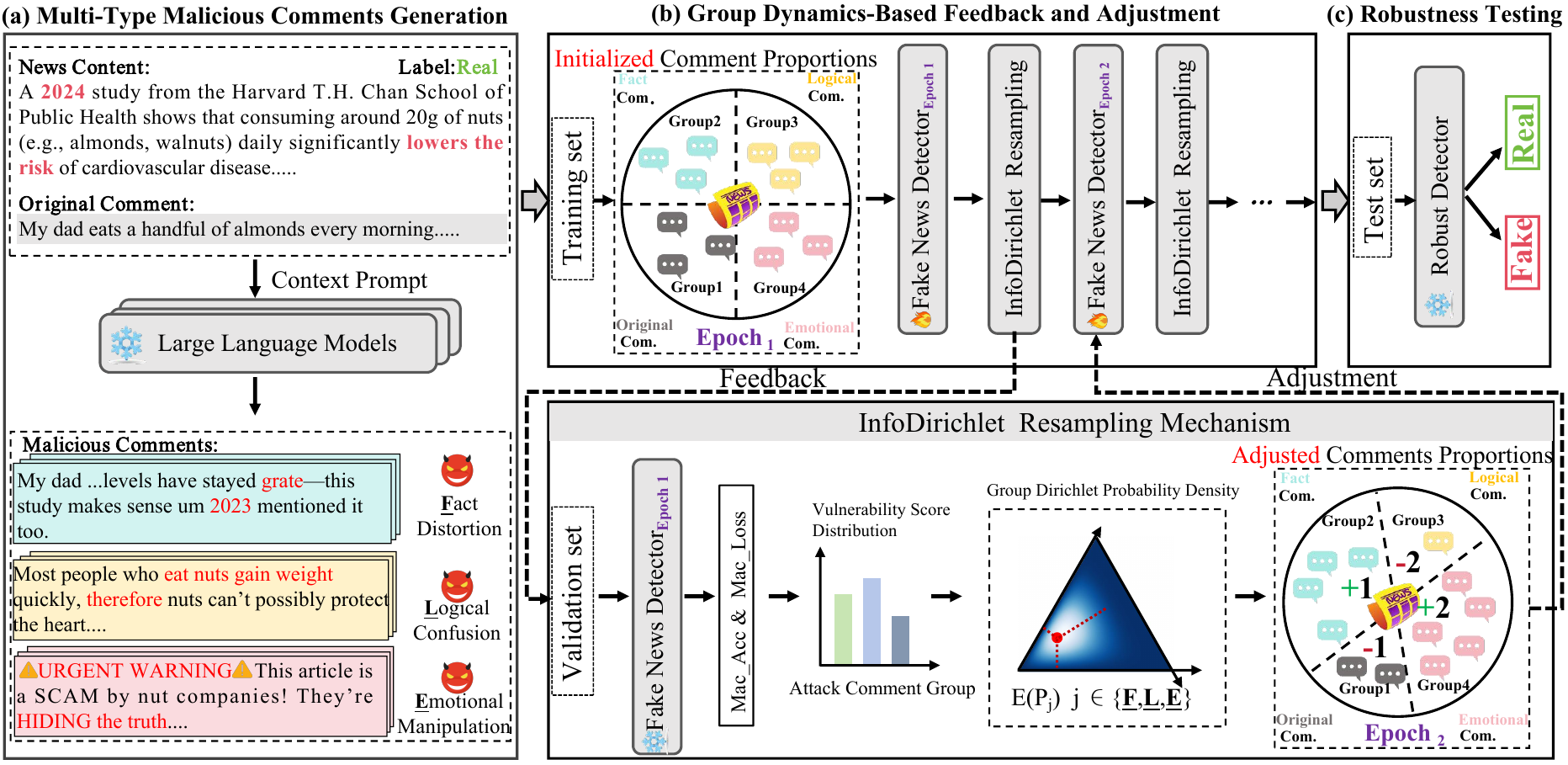}  
\caption{Overview of \textbf{CogniDir}. The framework consists of three stages: mechanism-specific malicious comment generation, feedback-driven sampling adjustment via InfoDirichlet Resampling, and robustness evaluation under mixed adversarial comment settings.}
    \label{fig:comprehensive_structure}
\end{figure*}

\section{Methodology}
In this section, we present \textbf{CogniDir}, a group-adaptive framework for robust fake news detection under malicious comment attacks. As illustrated in Fig.~\ref{fig:comprehensive_structure}, CogniDir first constructs mechanism-specific malicious comments with LLMs, covering fact distortion, logical confusion, and emotional manipulation. It then uses InfoDirichlet Resampling to estimate group-wise vulnerabilities after each training epoch and adaptively reallocate the sampling proportions of different malicious comment types. Finally, the trained detector is evaluated under mixed adversarial comment settings to assess its robustness against heterogeneous attack mechanisms.

\subsection{Multi-Type Malicious Comments Generation }
To better approximate the characteristics of malicious comments in real-world social environments, we categorize attackers' motivations into three complementary mechanisms based on cognitive psychology \citep{tversky1981framing,tversky1974judgment,tajfel1971social}: fact distortion, logical confusion, and emotional manipulation. However, existing public datasets provide only task-level annotations, such as news veracity or reply stance, without fine-grained annotations of cognitive manipulation mechanisms. Specifically, Weibo16 and Weibo20 provide news-level veracity labels, while RumourEval-19 additionally provides reply-level stance labels. None of these datasets contains cognitive-mechanism annotations, as detailed in Appendix~\ref{annotations}.

To address this gap, we use the strong reasoning capabilities of LLM to automatically generate malicious comments aligned with these mechanisms, yielding a dataset with explicit mechanism labels. Concretely, we design a prompt to simulate how real users produce malicious comments given a news article, while preserving linguistic naturalness and task relevance (see Appendix~\ref{app:data_quality} for dataset quality evaluation). Moreover, to reduce generator-specific style bias, we generate attacks using three models of different scales (Mistral-7B\citep{jiang2024mixtralexperts}, Gemma-2B\citep{team2024gemma}, and Qwen-32B\cite{hui2024qwen2}), and then perform stratified random sampling and combine them to form the final malicious comment dataset. We further examine potential generator-specific shortcuts by comparing single and mixed-generator settings, with the results reported in Appendix~\ref{app:generator_bias}. In summary, we design and generate a high-quality malicious comment dataset with diverse writing styles, enabling the robustness evaluation of malicious comment detection models against potential real-world malicious comments\citep{goodfellow2014explaining}.

\begin{figure}[!t]
\centering
\begingroup
\definecolor{promptTitleBg}{RGB}{210,220,235}
\definecolor{promptBodyBg}{RGB}{238,238,238}
\definecolor{promptFrame}{RGB}{120,120,120}

\newcommand{\rtext}[1]{\textcolor{red}{#1}}
\newcommand{\circnum}[1]{%
  \tikz[baseline=(X.base)] \node (X) [
    draw=black!80,
    circle,
    inner sep=0.6pt,
    line width=0.45pt,
    font=\normalsize
  ] {#1};%
}

\begin{tcolorbox}[
  width=0.45\textwidth,
  colback=promptBodyBg,
  colframe=promptFrame,
  colbacktitle=promptTitleBg,
  coltitle=black,
  title=\bfseries CoT Prompt for Attack Comment Generation,
  fontupper=\normalsize,
  boxrule=0.9pt,
  arc=0mm,
  left=2.6mm,right=2.6mm,
  top=1.4mm,bottom=1.4mm,
  toptitle=0.8mm,bottomtitle=0.8mm,
]
\setlength{\parindent}{0pt}
\setlength{\parskip}{0.15em}

\textbf{Task:} Given a news article and its original comment, generate an \rtext{attack comment}
that can \rtext{mislead authenticity judgment} while remaining content-relevant.

\textbf{Attack Type:} Fact Distortion \;|\; Logical Confusion\;|\; Emotional Manipulation.

\textbf{Two-step Strategy:} (1) extract \rtext{key points} and set a \rtext{target misleading direction};
(2) write the comment under the chosen attack type in a natural user tone.

\textbf{Output:} \emph{Key Points} + \emph{\rtext{Generated Attack Comment}}.
\end{tcolorbox}

\endgroup
\caption{Two-step CoT prompting strategy for generating attack comments with a specified attack type: understand the content first, then produce a realistic, content-grounded misleading comment.}
\label{fig:cot-prompt}
\vspace{-1em}
\end{figure}

\noindent\textbf{Prompt Design}. 
As shown in Fig.~\ref{fig:cot-prompt}, we adopt an \emph{understand-first, then-generate} CoT prompt to synthesize realistic malicious comments.
By first identifying salient content cues and then conditioning generation on them, the prompt improves content grounding and reduces off-topic or implausible attacks.
Meanwhile, specifying the attack type and requiring a natural user tone enables controllable generation across the three attack mechanisms.
To assess the realism and mechanism fidelity of the generated comments, we conduct a human evaluation on relevance, naturalness, attack-type consistency, and misleading potential, as shown in Appendix~\ref{app:human_eval}.
Further, to examine their real-world relevance, we qualitatively compare the generated attacks with real-world misleading user content exhibiting corresponding cognitive mechanisms, as shown in Appendix~\ref{app:real_world}.

\noindent \textbf{Dataset setting.} To support group-adaptive resampling, we organize the training, validation, and test sets according to attack mechanisms rather than only for static offline evaluation. Specifically, we aggregate all comment types, paired with their source news articles, into a unified training pool. Given a fixed budget of $M$ comments per instance, IDR dynamically controls the sampling proportions of different malicious groups from this pool, and we set $M=6$ according to the available generated comments. During training, group-specific validation sets, each containing comments from a single attack mechanism, are used to estimate attack-wise vulnerability and guide subsequent sampling updates. For testing, we uniformly sample from all malicious groups and combine them into mixed adversarial comment sets, enabling evaluation under multiple coexisting attack mechanisms.

\subsection{Group Dynamics-Based Feedback and Adjustment}
After getting a malicious comments combined datasets, we further need a robust detector to deal with the coexistence of multiple attack mechanisms. However, conventional methods typically train on all comments jointly without explicitly characterizing the detector’s differential vulnerabilities across attack types. As a result, training is dominated by an averaged objective, which often leads to imbalanced robustness across groups.

To address this, we introduce \emph{Group Dynamics-Based Feedback and Adjustment} in  CogniDir, a feedback-driven training module that consists of an \emph{InfoDirichlet Resampling} mechanism to compute group-wise vulnerability scores and adjust the proportions, thereby increasing the exposure of specific vulnerable group. By iterating \emph{Adaptive Resampling Training Loop}, the model can consistently focus on high vulnerability groups, gradually improving and balancing robustness across all groups.

\subsubsection{Detector Training}
At epoch $t$, we construct the comment set $\mathcal{C}_i^{(t)}$ for each news instance according to the current group-wise sampling distribution. The first epoch uses a uniform allocation over all attack groups, while subsequent epochs update this distribution based on the vulnerability scores estimated from validation feedback. Given the sampled training data, we optimize a shared detector $f_\theta$ by minimizing the empirical loss:
\begin{equation}
\theta^{\star}=\arg\min_{\theta}\ \sum_{i=1}^{N}\mathcal{L}\left(f_{\theta}(x_i,\mathcal{C}_i^{(t)}),y_i\right).
\end{equation}
Upon completing this training round, the model is specifically enhanced on the previous updated group distribution, thereby progressively improve robustness imbalance across different groups.

\subsubsection{InfoDirichlet Resampling}

To further assess the model’s robustness across different groups after this training round, we evaluate its vulnerability to different attack groups and adjust the training data distribution accordingly. Therefore, this section focuses on two key aspects: \one how to quantify group-wise vulnerability from validation signals; and \two how to convert these vulnerability measures into meaningful sampling proportions defined by a group probability distribution, enabling adaptive allocation of training resources across groups.

\noindent\textbf{\one Obtain Vulnerability Score.} After each training round, we evaluate the current detector on group-specific validation sets to get their performance. Notably, when faced with malicious comments, the detector needs to classify the article with multiple potentially contradictory comments, which often results in uncertain prediction.  Since accuracy alone cannot fully capture this phenomenon, we introduce probabilistic confidence as an additional metric to better characterize the model’s robustness against specific types of attacks \citep{guo2017calibrationmodernneuralnetworks}.

Concretely, we adopt an information-theoretic\citep{cover1991elements} view by focusing on the \emph{correctness event}. Let
$
Z=[\hat{y}=y]\in\{0,1\}
$
represent whether a prediction is correct. For each attack group $j\in\{F,L,E\}$, we evaluate Detector on the corresponding validation set $val_j$ (consisting only of samples from group $j$) and record the average cross-entropy loss and accuracy:
\begin{equation}
m_{j,\text{loss}},\ m_{j,\text{acc}}=\mathrm{Detector}(val_j),\qquad 
a_j\triangleq m_{j,\text{acc}}.
\end{equation}
Here, $a_j$ is the empirical frequency of the correctness event $Z$. We further convert the average negative log-likelihood into an equivalent probability mass,
\begin{equation}
q_j \triangleq \exp(-m_{j,\text{loss}}),
\end{equation}
which corresponds to the geometric-mean probability assigned to the true label (derivation in the Appendix~\ref{app:vuln_derivation}).

Based on the above, we define the group-wise vulnerability score, $s_j$, as the Bernoulli cross-entropy from the empirical parameter $a_j$ to the model-implied parameter $q_j$:
\begin{equation}
s_j \triangleq -a_j\log q_j-(1-a_j)\log(1-q_j).
\end{equation}
This admits the standard information-theoretic decomposition:
\begin{equation}
s_j = H(\mathrm{Bern}(a_j)) + D_{\mathrm{KL}}(\mathrm{Bern}(a_j)\,\|\,\mathrm{Bern}(q_j)),
\end{equation}
where the entropy term depends only on empirical correctness, while the KL divergence quantifies the mismatch between empirical correctness and the model-implied probabilistic characterization.

In short, this information-theoretic coupling of accuracy and confidence yields a group-wise vulnerability score $s_j$ for each attack group, which is subsequently used to derive the sampling/allocation probabilities in the next stage.

\vspace{1em}
\noindent\textbf{\two Adjust Comments Proportions.} After we obtain the vulnerability score $\{s_F, s_L, s_E\}$, we convert them into a group-level probability distribution to control the sampling budget across adversarial groups in subsequent training rounds. We aim to construct an allocation vector $p = (p_F, p_L, p_E)$ satisfying: (i) $p$ lies on the probability simplex (i.e., $p_j \ge 0$ and $\sum_j p_j = 1$), and (ii) $p$ varies smoothly with the relative magnitudes of $\{s_j\}$, enabling adaptive updates of the sampling strategy throughout training.

To obtain a group-probability mapping defined on the probability simplex, we model $p$ using a Dirichlet distribution\citep{bishop2006pattern}:
\begin{equation}
p \sim \mathrm{Dir}(\alpha), \qquad
f(p;\alpha)=\frac{1}{B(\alpha)}\prod_{j\in\{F,L,E\}}p_j^{\alpha_j-1},
\end{equation}
where $B(\alpha) = \frac{\prod_j \Gamma(\alpha_j)}{\Gamma(\sum_j \alpha_j)}$, and $\alpha = (\alpha_F, \alpha_L, \alpha_E)$ denotes the concentration parameters. Here, $\alpha$ serves as a bridge from vulnerability scores to a probability distribution: larger $\alpha_j$ leads to higher expected mass for group $j$, while the normalization inherent in the Dirichlet family ensures $\sum_j p_j = 1$.

We define $\alpha$ via a simple linear mapping from the vulnerability scores:
\begin{equation}
\alpha_j \triangleq s_j + 1, \qquad j \in \{F,L,E\},
\end{equation}
where the constant offset guarantees $\alpha_j > 0$, ensuring parameter validity. We adopt the mean of the Dirichlet distribution as a deterministic allocation strategy (see Appendix~\ref{app:allocation_theory}), which avoids the variance of random sampling and allows continuous updates with respect to $\{s_j\}$:
\begin{equation}
\bar{p}_j \triangleq \mathbb{E}[p_j] = \frac{\alpha_j}{\sum_{k \in \{F,L,E\}} \alpha_k}.
\end{equation}

This yields a smooth probability distribution $\bar{p}$ over the simplex, which is used to allocate the budget for malicious comments. Specifically, let $M$ denote the total number of comment set assignable to the three adversarial groups per instance. We allocate new comment set as follows:
\begin{equation}
\mathcal{C}_i^{(t+1)}
=\bigcup_{j\in\{F,L,E\}}
\mathrm{Sample}\!\left(\big\lfloor \bar{p}_j\, M \big\rfloor\right).
\end{equation}

In summary, Dirichlet Expectation Allocation provides a principled mechanism that smoothly maps vulnerability scores to group-level probability distributions, enabling stable and well-structured control over group-wise sampling throughout training. An equivalent entropy-regularized optimization interpretation and the
properties of this allocation are provided in Appendix~\ref{app:allocation_theory}.

\noindent\textbf{\three Adaptive Resampling Training Loop.} After applying InfoDirichlet Resampling to obtain a newly sampled comment set $\mathcal{C}_i^{(t+1)}$, we proceed to the next $t+1$ epoch of Detector Training.
In summary, this resampling step closes the feedback loop by reallocating training exposure toward the most vulnerable groups, enabling targeted improvement and more balanced robustness across attack types.

\subsection{Robustness Testing}
After the model converges at epoch \( T \), we apply the detector \( f_{\theta^\star} \) for classification. Given a test instance \((x_i, \mathcal{C}_i)\), we compute the prediction as:
\begin{equation}
\hat{y}_i = f_{\theta^\star}(x_i, \mathcal{C}_i'), \qquad \hat{y}_i \in \{0,1\}.
\end{equation}

In conclusion, by iteratively adjusting group-wise sampling proportions during training, we progressively reinforce the detector along its weakest attack groups and obtain a final model $f_{\theta^\star}$ that is robust and well-balanced across all malicious comment groups.

\section{Experiment}

In this section, we conduct comprehensive experiments to answer the following research questions:

\begin{itemize}
\item \textbf{RQ1: Attack Effectiveness.} How do malicious comments affect fake news detectors?
\item \textbf{RQ2: Robustness Enhancement.} Can CogniDir improve robustness against malicious comments?
\item \textbf{RQ3: Adaptive Allocation.} Does adaptive allocation reduce vulnerability gaps, and which components matter?
\end{itemize}

\subsection{Dataset} Considering for cultural and linguistic variation in information interpretation~\cite{nisbett2001culture}, we evaluate our model on the Chinese datasets Weibo16~\citep{ma2016detecting} and Weibo20~\citep{rao2021stanker}, and the English dataset RumourEval-19~\citep{gorrell2019semeval}. This cross-lingual setting enables a robust assessment under diverse social and discourse environments. Dataset statistics are summarized in Tab.~\ref{tab:dataset}.

\begin{table}[h]
\centering
\caption{Dataset statistics. pcs: number of news pieces; com: number of comments.}
\label{tab:dataset}
\resizebox{0.48\textwidth}{!}{%
\begin{tabular}{llrrrrrr}
\toprule
\multicolumn{2}{c}{} &
\multicolumn{2}{c}{Weibo-16} &
\multicolumn{2}{c}{Weibo-20} &
\multicolumn{2}{c}{RumourEval-19} \\
\cmidrule(lr){3-4}\cmidrule(lr){5-6}\cmidrule(lr){7-8}
Split & Veracity &
pcs & com &
pcs & com &
pcs & com \\
\midrule
\multirow{3}{*}{Training}
 & Fake  & 369 & 4,428  & 702  & 8,424  & 142 & 1,704 \\
 & Real  & 676 & 8,112  & 987  & 11,844 & 214 & 2,568 \\
 & Total & 1,045 & 12,540 & 1,689 & 20,268 & 356 & 4,272 \\
\midrule
\multirow{3}{*}{Validation}
 & Fake  & 154 & 924   & 318 & 1,908 & 50 & 600 \\
 & Real  & 316 & 1,896 & 421 & 2,526 & 82 & 984 \\
 & Total & 470 & 2,820 & 739 & 4,434 & 132 & 1,584 \\
\midrule
\multirow{3}{*}{Test}
 & Fake  & 70  & 420  & 119 & 714  & 16 & 96 \\
 & Real  & 112 & 672  & 177 & 1,062 & 39 & 234 \\
 & Total & 182 & 1,092 & 296 & 1,776 & 55 & 330 \\
\midrule
\multirow{3}{*}{All}
 & Fake  & 593  & 5,772  & 1,139 & 11,046 & 208 & 2,400 \\
 & Real  & 1,104 & 10,680 & 1,585 & 15,432 & 335 & 3,786 \\
 & Total & 1,697 & 16,452 & 2,724 & 26,478 & 543 & 6,186 \\
\bottomrule
\end{tabular}%
}
\end{table}

\begin{table*}[!h]
    \centering
    \tiny
    \caption{Performance comparison between \textbf{ CogniDir} and baselines on Weibo16, Weibo20, and RumourEval-19 datasets. "O" denotes original detection performance, "A" denotes performance following adversarial attack, and "R" denotes performance after robust training (with improvements over the attacked setting)—all reported metrics herein are F1 scores. The best performance is highlighted in bold. $\Delta$ \emph{Improve} represents our method's relative improvements over the second best performance in percentage.}
    \resizebox{1.0\textwidth}{!}{\rmfamily
        \begin{tabular}{llccccccccc}
            \toprule
            Type & Model
            & \multicolumn{3}{c}{Weibo16}
            & \multicolumn{3}{c}{Weibo20}
            & \multicolumn{3}{c}{RumourEval-19} \\
            \cmidrule(lr){3-5} \cmidrule(lr){6-8} \cmidrule(lr){9-11}
            & & O & A & R
              & O & A & R
              & O & A & R \\
            \midrule

            \multirow{4}{*}{LLM-Only}
            & Gemma-2-2B & 0.322 & 0.291 & 0.274 & 0.345 & 0.292 & 0.326 & 0.294 & 0.238 & 0.275 \\
            & Mistral-7B & 0.686 & 0.334 & 0.495 & 0.654 & 0.315 & 0.375 & 0.468 & 0.184 & 0.252 \\
            & Llama-3-8B & 0.598 & 0.324 & 0.408 & 0.582 & 0.363 & 0.408 & 0.377 & 0.219 & 0.276 \\
            & Qwen2.5-32B & 0.831 & 0.282 & 0.716 & 0.823 & 0.367 & 0.541 & 0.553 & 0.343 & 0.412 \\
            \midrule

            \multirow{5}{*}{Deep-Learning Based}
            & dEFEND & 0.902 & 0.727 & 0.745 & 0.884 & 0.557 & 0.585 & 0.648 & 0.407 & 0.687 \\
            & Dual-CAN & 0.895 &  0.366 & 0.578 & 0.872 & 0.436 & 0.580 & 0.628 &  0.377 & 0.546 \\
            & GenFEND & 0.915 & 0.578 & 0.867 & 0.892 & 0.640 & 0.761 & 0.662 & 0.456 & 0.606 \\
            & L-Defense & 0.882 & 0.706 & 0.759 & 0.860 & 0.666 & 0.702 & 0.610 & 0.343 & 0.709 \\
            & ARG & 0.884 & 0.619 & 0.820 & 0.812 & 0.576 & 0.803 & 0.712 & 0.440 & 0.609 \\
            &  CogniDir & \textbf{0.955} & \textbf{0.792} & \textbf{0.945} & \textbf{0.936} & \textbf{0.728} & \textbf{0.947} & \textbf{0.822} & \textbf{0.552} & \textbf{0.812} \\
            \midrule

            \rowcolor{gray!15}
            \multicolumn{1}{c}{} & $\mathit{\Delta Improve}$
            & +4.4\% & +8.9\% & +9.0\%
            & +4.9\% & +9.3\% & +17.9\%
            & +15.5\% & +21.1\% & +14.5\% \\
            
            \bottomrule
        \end{tabular}
    }
    \label{tab:main_rewrite}
\end{table*}

\vspace{-1.5em}
\subsection{Implementation Details}

\noindent \textbf{Model Architecture.} For the Detector, we adopt a comment-aware classifier. We use a shared pretrained BERT encoder to extract representations for the news article and each associated comment. These representations are refined by a self-attention layer and be concatenated; finally, an MLP classifier outputs the binary classification result.

\noindent \textbf{Attack Settings.} To systematically assess robustness under diverse adversarial behaviors, we construct two complementary evaluation settings: \emph{group-specific} and \emph{mixed}. Group-specific sets are formed by composing each test instance exclusively with comments from a single attack type, enabling targeted diagnosis of attack-wise failure modes and worst-case susceptibility. In contrast, the mixed set integrates multiple attack types within each instance to approximate realistic comment streams and measure robustness under combined adversarial interference. In conclusion, these two settings jointly capture both per-attack brittleness and practical robustness under real-world attack mixtures. Throughout evaluation, O denotes performance on the original comment set, A denotes performance on the attacked comment set before robust training, and R denotes performance on the same attacked test inputs after robust training.

\vspace{3mm}

\begin{figure*}[!h]
  \centering
  \begin{minipage}[t]{0.48\textwidth}
    \centering
    \includegraphics[width=\linewidth]{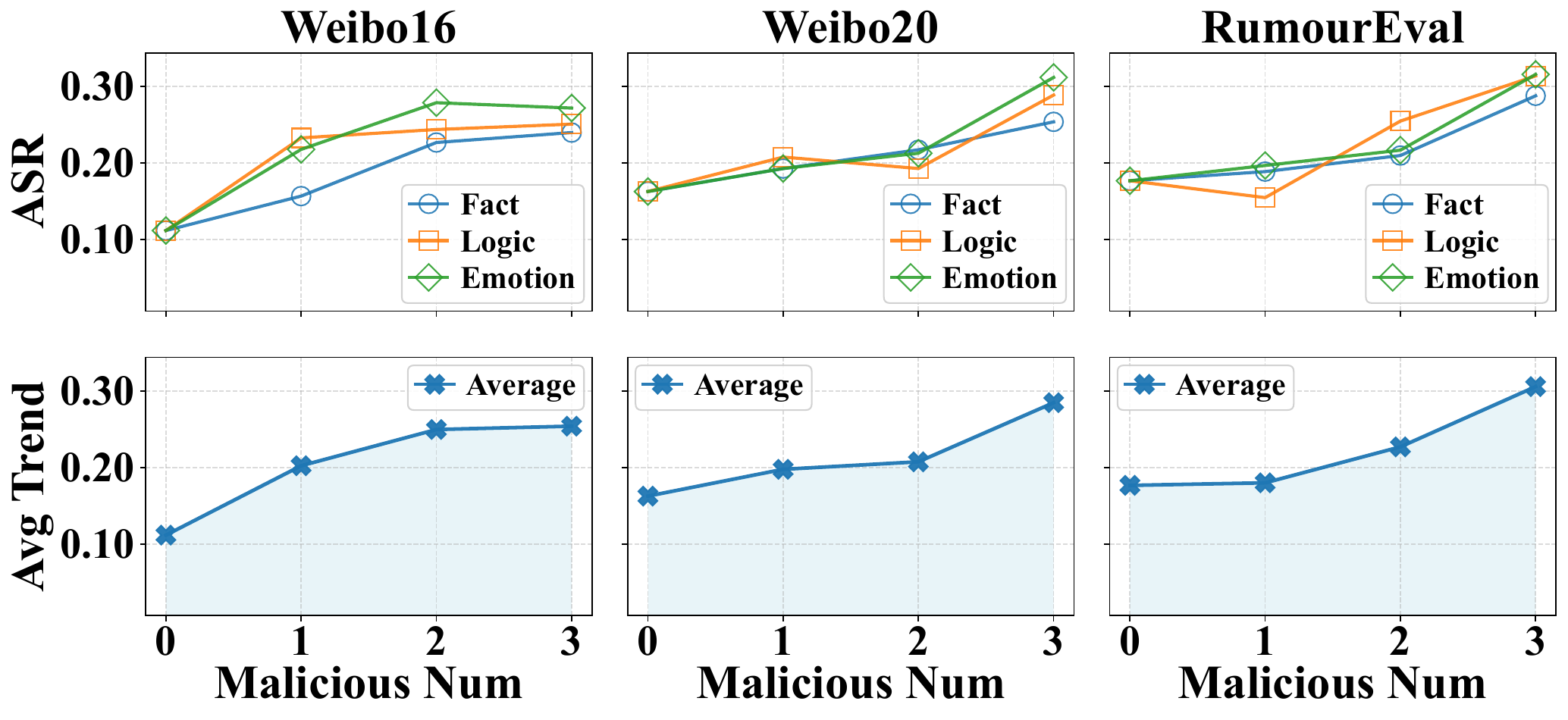}
    \caption{ASR \textbf{before} robust training under varying attack types and numbers. Top: Fact/Logic/Emotion ASR under the number of injected malicious comments across three datasets. Bottom: mean ASR averaged over attack types at each attack number.}
    \label{fig:asr_ori}
  \end{minipage}\hfill
  \begin{minipage}[t]{0.48\textwidth}
    \centering
    \includegraphics[width=\linewidth]{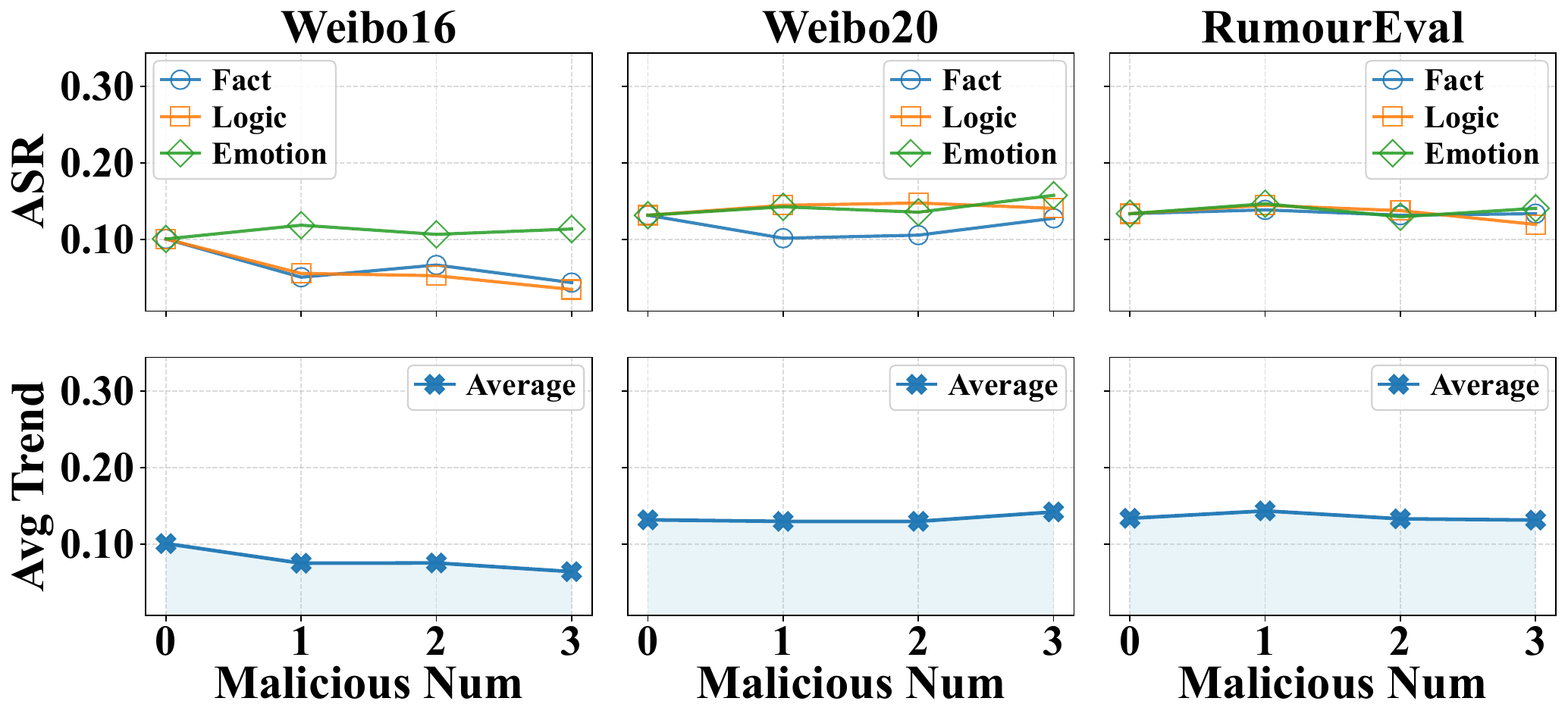}
    \caption{ASR \textbf{after} robust training under varying attack types and numbers. Top: Fact/Logic/Emotion ASR under the number of injected malicious comments across three datasets. Bottom: mean ASR averaged over attack types at each attack number.}
    \label{fig:asr_com}
  \end{minipage}
  \vspace{-3mm}
\end{figure*}
\subsection{Baselines}

We select representative baselines from two categories: (i) \textbf{LLM-Only}:  We use four LLMs of different sizes: Gemma-2-2B \citep{team2024gemma}, Mistral-7B \citep{jiang2023mistral}, Llama-3-8B\citep{grattafiori2024llama}, and Qwen2.5-32B\cite{hui2024qwen2}).
(ii) \textbf{Deep-Learning Based}: we adopt five representative deep-learning fake news detectors dEFEND\citep{shu2019defend}, Dual-CAN \citep{yang2023entity}, GenFEND \citep{nan2024let}, L-Defense \citep{wang2024explainable}, ARG \citep{hu2024bad}, and our proposed \textbf{ CogniDir}. Details of the baselines and evaluation protocols are provided in the Appendix~\ref{app:baselines}.

\textbf{Robustness Testing Analysis.}
CogniDir exhibits robustness under malicious comments, mitigating performance degradation caused by malicious inputs. Traditional detectors suffer an average performance drop of over 18\% under different attack types, whereas CogniDir narrows this gap, maintaining an F1 above 0.94 across attack categories. This robustness stems from our adaptive training strategy, which dynamically reallocates learning focus toward more vulnerable attack groups via the InfoDirichlet Resampling. To assess result stability, we additionally repeat CogniDir and GenFEND with three matched random seeds. Detailed mean$\pm$standard deviation and paired significance tests are reported in Appendix~\ref{app:multi-run}.

\textbf{LLM-Only vs. Deep-Learning Based Analysis.}
When comparing LLM-only and Deep-Learning Based detectors, we observe that LLMs such as Qwen2.5-32B perform competitively in non-adversarial settings but exhibit pronounced vulnerability under attack, e.g., F1 on Weibo16 drops sharply from 0.831 to 0.282. In contrast, deep-learning models like dEFEND and GenFEND show stronger baseline robustness yet still leave considerable room for improvement.  CogniDir bridges this gap by integrating a deep-learning architecture with LLM-generated adversarial examples and adaptive training. It leverages LLMs to synthesize diverse and realistic attack samples, while employing a dynamic training mechanism to harden the detector. This combined strategy allows  CogniDir to achieve both high initial accuracy and sustained robustness, outperforming all LLM-only and deep-learning baselines in adversarial environments.

\begin{tcolorbox}[
colback=blue!3!white,colframe=blue!20!gray,
boxrule=0.6pt,arc=1mm,left=1.5mm,right=1.5mm,
top=0.8mm,bottom=0.8mm]
\small
\textbf{RQ1--RQ2.}
Malicious comments consistently degrade existing detectors, while CogniDir outperforms the strongest baselines by \textbf{8.9--21.1\%} under attack.
\end{tcolorbox}

\subsection{Robust Analysis}

We evaluate the robustness of  CogniDir under different malicious comment numbers on Weibo16, Weibo20, and RumourEval-19. We first measure attack success rate(ASR) without adversarial training (as shown Fig. \ref{fig:asr_ori}), then assess improvements after incorporating adversarial samples during training (as shown Fig. \ref{fig:asr_com}), analyzing how robustness scales with attack intensity across categories.

\noindent \textbf{Effect of malicious comment numbers.}  
When increasing the number of malicious comments from 0 to 3 in Fig.\ref{fig:asr_ori}, the ASR across datasets rises from 0.145 to 0.287, an increase of nearly 98\%. This trend confirms that denser adversarial inputs accumulate semantic noise and contradictory cues, which destabilize the model’s feature representation and decision boundary. Hence, higher attack intensity amplifies uncertainty and weakens detection reliability across all attack types.

\noindent \textbf{Effect of Malicious Comment Types.}  
As shown in Fig.~\ref{fig:asr_ori}, the average Attack Success Rates (ASR) are $0.206$ (fact distortion), $0.241$ (logical confusion), and $0.283$ (emotional manipulation). This indicates the model is most robust to fact distortion, whose lexical noise is contextually mitigated; moderately affected by logical confusion; and most vulnerable to emotional manipulation, which alters sentiment to severely mislead authenticity judgment.

\noindent \textbf{Effect of Robust Training.}  
As shown in Fig.\ref{fig:asr_com}, after incorporating adversarial group training, the mean ASR drops to 0.116, representing an overall 45\% reduction compared with the untrained setting. Moreover, inter-attack variance decreases sharply (from over 0.07 to below 0.03), indicating that the model becomes both more accurate and more stable across attack types and densities. These findings suggest that our robust training strategy enables the model to learn generalized and invariant representations, maintaining consistent detection performance even under complex adversarial scenarios.

\begin{tcolorbox}[
colback=blue!3!white,colframe=blue!20!gray,
boxrule=0.6pt,arc=1mm,left=1.5mm,right=1.5mm,
top=0.8mm,bottom=0.8mm]
\small
\textbf{RQ2.}
CogniDir remains robust across attack types and different numbers of malicious comments, reducing the mean ASR by \textbf{45\%}.
\end{tcolorbox}

\subsection{Convergence Analysis}

\begin{figure}[h]
  \centering
  \includegraphics[width=0.475\textwidth]{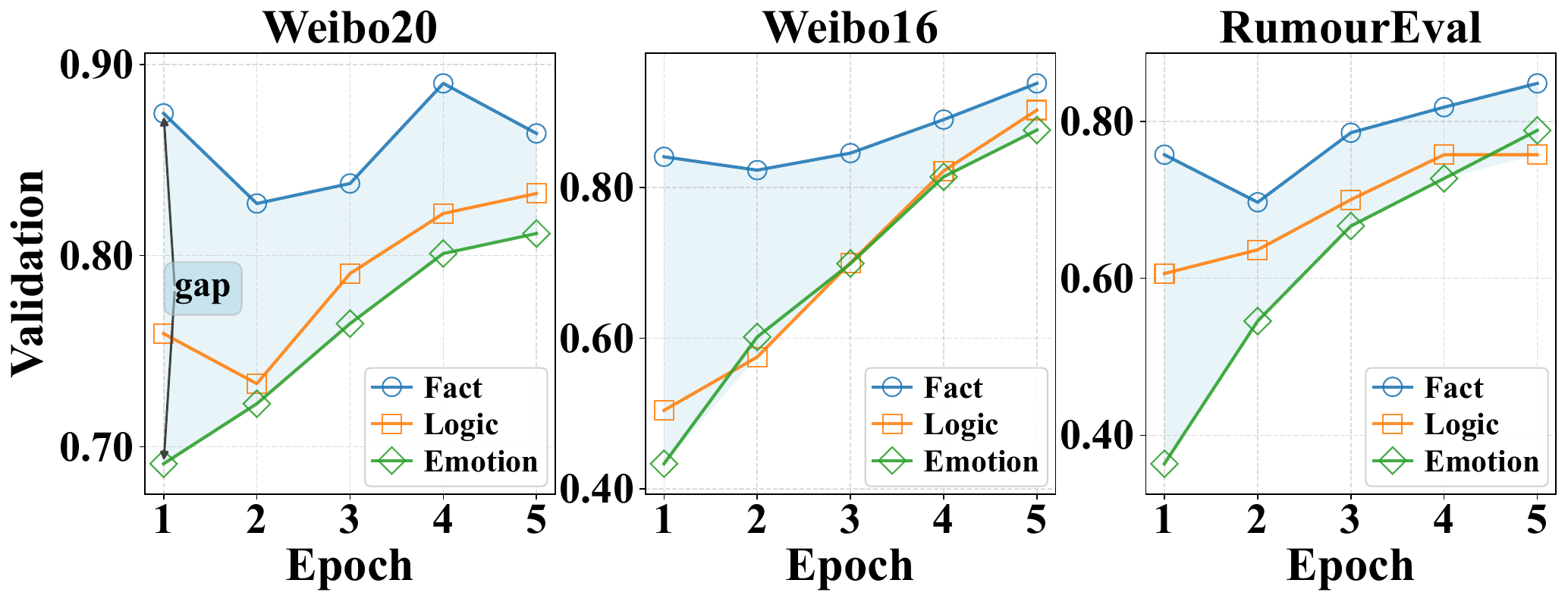}
  \caption{Validation accuracy across epochs on the three attack-type validation sets.}
  \label{fig:Convergence_Fig}
\end{figure}

To empirically verify the adaptive robustness improvement across different attack categories during the training process, we continuously track the validation accuracy on three distinct adversarial groups: Fact Distortion, Logical Confusion, and Emotional Manipulation. As illustrated in Fig.~\ref{fig:Convergence_Fig}, the model exhibits significant and consistent performance gains across all categories. Specifically, the mean accuracy rises from 0.82 to 0.90 for Fact Distortion, from 0.69 to 0.83 for Logical Confusion, and shows the most substantial improvement for Emotional Manipulation, increasing from an initial 0.59 to 0.82. More importantly, the performance disparity among these groups, measured by the inter-group gap, dramatically narrows from 0.23 to 0.08. This convergence indicates that InfoDirichlet Resample effectively prevents the model from overfitting to easier attack types. Instead, it progressively reallocates the learning focus toward the more challenging, weaker groups, ultimately yielding a more balanced and stable robustness against diverse and complex adversarial comment mechanisms.

\begin{tcolorbox}[
colback=blue!3!white,colframe=blue!20!gray,
boxrule=0.6pt,arc=1mm,left=1.5mm,right=1.5mm,
top=0.8mm,bottom=0.8mm]
\small
\textbf{RQ3.}
Adaptive allocation progressively balances robustness across heterogeneous attack groups.
\end{tcolorbox}

\vspace{-3mm}
\subsection{Ablation Study}
To quantify the contribution of each component in  CogniDir, we conduct an ablation study on Malicious Comments Generation (G), Vulnerability Score (VS), and Dirichlet Expectation Allocation (DEA). Based on Tab.~\ref{tab:ablation}, we show the model's performance across the three datasets when removing a single module. Compared with the full model, removing G ( CogniDir-G) yields the largest performance drop, with an average macro-F1 decrease of 21.4\% points. Replacing VS with random scores ( CogniDir-VS) results in an average decrease of 16.3\% points, while substituting DEA with a softmax-weighted scheme ( CogniDir-DEA) also leads to a clear degradation, reducing macro-F1 by 11.1\% points on average. These results indicate that the three components are essential and complementary for robust adversarial training, and removing any of them substantially weakens overall performance.

\begin{tcolorbox}[
colback=blue!3!white,colframe=blue!20!gray,
boxrule=0.6pt,arc=1mm,left=1.5mm,right=1.5mm,
top=0.8mm,bottom=0.8mm]
\small
\textbf{RQ3.}
Generation, vulnerability estimation, and adaptive allocation are complementary and all materially contribute to robustness.
\end{tcolorbox}

\vspace{-2mm}
\begin{table}[h]
\centering
\caption{Evaluation results of ablating G, VS and DEA from  CogniDir on both datasets.}
\label{tab:ablation}
\resizebox{\linewidth}{!}{
\begin{tabular}{l|l c c c c}
\hline
Dataset & Method & mac F1 & Acc. & F1-real & F1-fake \\
\hline
\multirow{5}{*}{Weibo16} 
                          &  CogniDir-G   & 0.792 & 0.811 & 0.830 & 0.754\\
                          &  CogniDir-VS & 0.821 & 0.836 & 0.862 & 0.780 \\
                          &  CogniDir-DEA   & 0.855 & 0.872 & 0.895 & 0.815 \\
                          &  CogniDir     & \textbf{0.945} & \textbf{0.947} & \textbf{0.957} & \textbf{0.933} \\
\hline
\multirow{5}{*}{Weibo20} 
                          &  CogniDir-G   & 0.728 & 0.742 & 0.770 & 0.686 \\
                          &  CogniDir-VS & 0.792 & 0.807 & 0.825 & 0.759 \\
                          &  CogniDir-DEA   & 0.834 & 0.848 & 0.872 & 0.796 \\
                          &  CogniDir     & \textbf{0.947} & \textbf{0.947} & \textbf{0.951} & \textbf{0.943} \\
\hline
\multirow{5}{*}{RumourEval-19} 
                                &  CogniDir-G   & 0.552 & 0.574 & 0.620 & 0.484 \\
                                &  CogniDir-VS & 0.613 & 0.628 & 0.662 & 0.564 \\
                                &  CogniDir-DEA   & 0.692 & 0.707 & 0.748 & 0.636 \\
                                &  CogniDir     & \textbf{0.812} & \textbf{0.836} & \textbf{0.883} & \textbf{0.741} \\
\hline
\end{tabular}
}
\end{table}
\vspace{-6mm}
\section{Conclusion}  

In this study, we design and evaluate three categories of malicious comments grounded in cognitive psychology: fact distortion, logical confusion, and emotional manipulation. Based on these mechanisms, we further propose CogniDir, which dynamically estimates group-wise vulnerability and adjusts training exposure toward more vulnerable attack types. Experimental results across three benchmark datasets demonstrate that CogniDir achieves strong robustness and detection performance under heterogeneous malicious-comment attacks. Further analyses show that adaptive allocation progressively reduces the vulnerability gap across attack groups, while the ablation study confirms the complementary contributions of malicious-comment generation, vulnerability estimation, and adaptive allocation. These findings indicate that explicitly modeling heterogeneous attack mechanisms is important for robust comment-aware fake news detection. Overall, CogniDir provides a practical framework for mechanism-aware and feedback-driven robust training in evolving adversarial social media environments.

\clearpage

\section*{Limitations}

Despite the effectiveness of our proposed CogniDir, there are several limitations. First, our current evaluation primarily focuses on mechanism-controlled malicious comments synthesized by multiple LLMs. Although mixed-generator synthesis and human evaluation improve the diversity and quality of these comments, they cannot fully represent naturally occurring malicious comments in real-world social media. Second, we do not exhaustively evaluate attacks generated by unseen LLMs or unseen prompting strategies. Future work will extend CogniDir to naturally occurring and out-of-distribution comment attacks to further assess its robustness under more diverse adversarial settings.

\section*{Acknowledgements}
This work was supported by the National Natural Science Foundation of China under Grants 62376265, 62502514, and 62372454.

\bibliography{custom}

\appendix
\clearpage

\section*{Appendix Overview}

The appendix is organized as follows:

\begin{itemize}
    \item \textbf{Appendix A} Threat model and direction-specific attack analysis.
    \item \textbf{Appendix B} Original dataset annotation schemes.
    \item \textbf{Appendix C} Generated comment construction and quality analysis.
    \item \textbf{Appendix D} Generator selection and bias analysis.
    \item \textbf{Appendix E} Human evaluation.
    \item \textbf{Appendix F} Real-world mechanism correspondence.
    \item \textbf{Appendix G} Derivation for vulnerability score.
    \item \textbf{Appendix H} Theoretical view of adaptive allocation.
    \item \textbf{Appendix I} Baselines and evaluation protocols.
    \item \textbf{Appendix J} Statistical analysis across multiple runs.
\end{itemize}

\section{Threat Model and Direction-Specific Attack Analysis}
\label{app:directional}

We consider an untargeted prediction-manipulation attack against comment-aware fake news detectors. The attacker keeps the news content unchanged and injects malicious comments into the associated comment set. An attack succeeds whenever an originally correct prediction becomes incorrect, without specifying a target label in advance.

Accordingly, the overall attack success rate includes both real-to-fake and fake-to-real prediction flips. For the two directions, we define

\begin{equation}
\mathrm{ASR}_{R\rightarrow F}
=
\frac{
\sum_i
\mathbb{I}
\left[
y_i=0
\wedge
\hat{y}_i^{O}=0
\wedge
\hat{y}_i^{A}=1
\right]
}{
\sum_i
\mathbb{I}
\left[
y_i=0
\wedge
\hat{y}_i^{O}=0
\right]
},
\end{equation}

and

\begin{equation}
\mathrm{ASR}_{F\rightarrow R}
=
\frac{
\sum_i
\mathbb{I}
\left[
y_i=1
\wedge
\hat{y}_i^{O}=1
\wedge
\hat{y}_i^{A}=0
\right]
}{
\sum_i
\mathbb{I}
\left[
y_i=1
\wedge
\hat{y}_i^{O}=1
\right]
}.
\end{equation}

Using GenFEND as a representative baseline, Tab.~\ref{tab:directional_asr} reports the attack effectiveness in both directions.

\begin{table}[t]
\centering
\small
\setlength{\tabcolsep}{4pt}
\caption{Direction-specific attack success rates(ASR) on GenFEND.}
\label{tab:directional_asr}
\resizebox{\columnwidth}{!}{
\begin{tabular}{lccc}
\toprule
Dataset & F1 under Attack & $R\rightarrow F$  & $F\rightarrow R$  \\
\midrule
Weibo16       & 0.578 & 36.4\% & 47.9\% \\
Weibo20       & 0.640 & 39.0\% & 31.6\% \\
RumourEval-19 & 0.456 & 54.5\% & 50.0\% \\
\bottomrule
\end{tabular}
}
\end{table}

The substantial attack success rates in both directions indicate that the observed degradation reflects a general vulnerability to malicious comment manipulation instead of an attack specialized toward a single target label.

\section{Original Dataset Annotation Schemes}
\label{annotations}

The original datasets primarily provide task-level annotations for fake news detection. Weibo16 and Weibo20 provide news-level veracity labels, while RumourEval-19 additionally provides reply-level stance labels (as shown in Tab.\ref{tab:dataset_annotations}). None of the three datasets contains annotations for the cognitive manipulation mechanisms studied in this work.

\begin{table}[t]
\centering

\setlength{\tabcolsep}{4pt}
\caption{Annotation schemes provided by the original datasets.}
\label{tab:dataset_annotations}
\resizebox{\columnwidth}{!}{
\begin{tabular}{lccc}
\toprule
Dataset & Veracity & Comment Label & Mechanism Label \\
\midrule
Weibo16       & Yes & None   & None \\
Weibo20       & Yes & None   & None \\
RumourEval-19 & Yes & Stance & None \\
\bottomrule
\end{tabular}
}
\end{table}

Consequently, the original annotations cannot directly characterize the cognitive manipulation embodied in malicious comments or distinguish the effects of different attack mechanisms.

\section{Generated Comment Construction and Quality}
\label{app:data_quality}

\subsection{Prompt Construction Protocol}
\label{app:prompt_protocol}

The attack-generation prompt is derived from the task definition rather than selected according to downstream detection performance. It follows an understand-first, then-generate procedure. The first stage identifies salient facts in the news and determines a misleading direction consistent with the specified attack mechanism. The second stage generates a comment conditioned on the news context, attack mechanism, and natural user tone.

The prompt template was fixed before large-scale generation and applied consistently across datasets and generation models. No prompt variants were selected based on downstream detection results.

\subsection{Lexical Diversity via Distinct-$n$}
\label{app:distinct}

We compute Distinct-$n$ for $n\in\{1,2,3\}$ on the generated comments to examine lexical diversity.

\begin{table}[t]
\centering
\small
\setlength{\tabcolsep}{4pt}
\caption{Lexical diversity of generated malicious comments.}
\label{tab:distinct}
\begin{tabular}{lccc}
\toprule
Dataset & Distinct-1 & Distinct-2 & Distinct-3 \\
\midrule
Weibo16       & 0.29 & 0.58 & 0.76 \\
Weibo20       & 0.31 & 0.60 & 0.78 \\
RumourEval-19 & 0.27 & 0.55 & 0.73 \\
\bottomrule
\end{tabular}
\end{table}

\subsection{Standardized OOD Distance}
\label{app:ood}

We further compare generated comments with original comments in representation space. Given a fixed multilingual sentence encoder $\phi$, we compute the centroid $\mu_{\mathrm{real}}$ of the original comments. The cosine distance of a comment $c$ from this centroid is

\begin{equation}
d(c)
=
1-\cos\left(\phi(c),\mu_{\mathrm{real}}\right).
\end{equation}

Using the mean $\mu_d$ and standard deviation $\sigma_d$ of the distances of original comments, we define the standardized distance as

\begin{equation}
d_{\mathrm{std}}(c)
=
\frac{d(c)-\mu_d}{\sigma_d}.
\end{equation}

The mean standardized distances are 0.25, 0.32, and 0.35 for Fact Distortion, Logical Confusion, and Emotional Manipulation, respectively.

\subsection{Case Study: Generated Adversarial Comments}
\label{app:generated_cases}

To illustrate the distinctions among the three attack mechanisms, we present representative generated comments conditioned on the same news example.

\begin{exampleboxer}{News Sample (Label: Real)}
\textbf{News Content:}
``Local health bureau: No new respiratory illness cases this week. All patients recovered. Continue wearing masks in crowded areas.''
\end{exampleboxer}

\begin{examplebox}{Fact Distortion}
\textbf{Generated Comment:}
``I checked the hospital WeChat---they actually reported 47 new cases last week but only counted recovered ones to keep the number low!''

\smallskip
\emph{\textbf{Why misleading:}}
The comment fabricates factual details and numerical evidence, creating false certainty and undermining trust in the reported statistics.
\end{examplebox}

\begin{examplebox}{Logical Confusion}
\textbf{Generated Comment:}
``If there are really zero new cases, why are they still telling us to wear masks everywhere? Either the treatment doesn't work or the data is fake.''

\smallskip
\emph{\textbf{Why misleading:}}
The comment introduces flawed causal reasoning and a false dichotomy that creates doubt while appearing superficially coherent.
\end{examplebox}

\begin{examplebox}{Emotional Manipulation}
\textbf{Generated Comment:}
``My friend in the hospital says they're hiding deaths to avoid panic before the holiday... Don't believe the official numbers, protect your family!''

\smallskip
\emph{\textbf{Why misleading:}}
The comment uses fear, conspiracy framing, and personal anecdotes to encourage authenticity judgments driven by emotion instead of verifiable evidence.
\end{examplebox}

\section{Generator Selection and Bias Analysis}
\label{app:generator_bias}

The main experiments combine malicious comments generated by Mistral-7B, Gemma-2B, and Qwen-32B. This mixed-generator design is intended to reduce the possibility that the detector learns generator-specific vocabulary, sentence patterns, or semantic styles.

To examine this issue, we compare the mixed-generator setting with a single-generator setting in which Qwen-32B is used for both robust-training and test attack comments.

\begin{table}[t]
\centering
\small
\setlength{\tabcolsep}{4pt}
\caption{Macro-F1 under different attack-generator settings.}
\label{tab:generator_bias}
\resizebox{\columnwidth}{!}{
\begin{tabular}{lccc}
\toprule
Dataset & Original & Qwen-32B Robust & Mixed Robust \\
\midrule
Weibo16       & 0.955 & 0.948 & 0.945 \\
Weibo20       & 0.936 & 0.951 & 0.947 \\
RumourEval-19 & 0.822 & \textbf{0.924} & 0.812 \\
\bottomrule
\end{tabular}
}
\end{table}

The single-generator setting produces an unusually large result on RumourEval-19, where robust performance reaches 0.924 compared with an original performance of 0.822. Because the same generator is used for training and testing in this setting, this improvement may reflect generator-specific shortcuts in addition to genuine robustness recovery. The mixed-generator setting does not exhibit the same anomalous increase. We therefore use mixed generation in the main experiments to reduce dependence on the stylistic characteristics of a single generator.

\section{Human Evaluation}
\label{app:human_eval}

We conduct a stratified human evaluation of the generated malicious comments. Samples are stratified by dataset (\textsc{Weibo16}, \textsc{Weibo20}, and \textsc{RumourEval-19}), veracity label (real or fake), and attack mechanism (Fact Distortion, Logical Confusion, and Emotional Manipulation), yielding 18 strata. We uniformly sample 30 comments from each stratum, resulting in $N=540$ evaluated comments.

Each comment is independently assessed by two graduate-level annotators according to four criteria:

\begin{itemize}
    \item \textbf{Relevance:} The comment is on-topic and refers to the entities or events described in the news.
    \item \textbf{Naturalness:} The comment resembles plausible user-generated content in tone and expression.
    \item \textbf{Attack-Type Fidelity:} The comment is consistent with the intended cognitive attack mechanism.
    \item \textbf{Misleading Potential:} The comment can plausibly interfere with authenticity judgment.
\end{itemize}

A sample passes the quality check when it satisfies relevance, naturalness, and attack-type fidelity. Disagreements are resolved through discussion or additional adjudication. Inter-annotator agreement on acceptance is substantial, with Cohen's $\kappa=0.72$.

\begin{table}[t]
\centering
\small
\setlength{\tabcolsep}{6pt}
\caption{Human-evaluation acceptance rates.}
\label{tab:human_eval}
\begin{tabular}{lc}
\toprule
Attack Mechanism & Acceptance Rate \\
\midrule
Fact Distortion         & 92.1\% \\
Logical Confusion       & 90.3\% \\
Emotional Manipulation  & 89.4\% \\
\midrule
Overall                 & 90.6\% \\
\bottomrule
\end{tabular}
\end{table}

The high acceptance rates across all three mechanisms support the linguistic quality and mechanism fidelity of the generated comments. Rejected samples are regenerated once.

\section{Real-World Mechanism Correspondence}
\label{app:real_world}

To examine whether the three mechanisms correspond to patterns observed in real-world misleading user content, we qualitatively compare generated attack comments with representative real-world cases.

\begin{table*}[t]
\centering
\scriptsize
\caption{Qualitative correspondence between real-world misleading content and generated attack comments.}
\label{tab:real_world_cases}
\begin{tabular}{p{0.18\textwidth}p{0.24\textwidth}p{0.25\textwidth}p{0.25\textwidth}}
\toprule
Case & Real-World Misleading Content & Generated Attack Comment & Mechanism Correspondence \\
\midrule

``Xu Jing'' family-search post during the Ya'an Lushan earthquake
&
A reposted help message used a specific name, hospital, location, and contact information to create credibility and encourage further dissemination.
&
``This urgent request for help appears highly credible and provides specific details, including a name, location, and phone number. I hope Xu Jing receives the message soon. Please help spread it!''
&
\textbf{Fact Distortion.} Both rely on specific but unverified details to create false credibility and encourage dissemination.
\\[2pt]

Rapid increase in signatures on a South Korean Blue House national petition
&
A rapid increase in comments and petition recommendations was interpreted as evidence that unspecified external forces were manipulating the process.
&
``The surging number of petition signatures raises doubts about its authenticity. Receiving more than 17,000 signatures overnight is usually associated with data manipulation.''
&
\textbf{Logical Confusion.} Both infer deliberate manipulation directly from a short-term increase in quantity without intermediate evidence.
\\[2pt]

``Crisis actor'' narrative surrounding the Boston Marathon bombing
&
Inconsistent footage and eyewitness accounts were framed as evidence that the event had been staged.
&
``The victims in the incident were actually U.S. soldiers who lost limbs in Afghanistan, and the scene appears to have been carefully staged, which deserves further thought.''
&
\textbf{Emotional Manipulation.} Both use conspiratorial narratives to induce shock and distrust and to shift judgment away from verifiable evidence.
\\

\bottomrule
\end{tabular}
\end{table*}

These examples illustrate qualitative correspondence between the mechanisms instantiated by our generated comments and misleading patterns observed in real-world user content. They do not constitute a representative benchmark of naturally occurring attacks, which remains an important direction for future evaluation.

\section{Derivation for Vulnerability Score}
\label{app:vuln_derivation}

For attack group $j$, let

\begin{equation}
\hat{p}_i
=
p_\theta(y_i\mid x_i,\mathcal{C}_i)
\end{equation}

denote the probability assigned to the true label for validation sample $i$. The average cross-entropy loss is

\begin{equation}
m_{j,\mathrm{loss}}
=
-\frac{1}{n}
\sum_{i=1}^{n}
\log \hat{p}_i.
\end{equation}

The corresponding confidence proxy is

\begin{equation}
q_j
=
\exp(-m_{j,\mathrm{loss}})
=
\left(
\prod_{i=1}^{n}
\hat{p}_i
\right)^{1/n},
\end{equation}

which is the geometric mean of the true-label probabilities.

Let

\begin{equation}
a_j = m_{j,\mathrm{acc}}
\end{equation}

denote empirical accuracy for attack group $j$. We define the group-wise vulnerability score as

\begin{equation}
s_j
=
-a_j\log q_j
-
(1-a_j)\log(1-q_j).
\end{equation}

This score can be decomposed as

\begin{equation}
s_j
=
H(\mathrm{Bern}(a_j))
+
D_{\mathrm{KL}}
\left(
\mathrm{Bern}(a_j)
\|
\mathrm{Bern}(q_j)
\right).
\end{equation}

The entropy term characterizes uncertainty in empirical correctness, while the KL term captures the mismatch between empirical correctness and the model-implied true-label confidence. The score therefore couples accuracy and confidence in a fixed form and does not introduce a manually tuned combination weight.

\section{Theoretical View of Adaptive Allocation}
\label{app:allocation_theory}

\subsection{Dirichlet-Mean Parameterization}

For each attack group $j\in\{F,L,E\}$, we map the vulnerability score to a positive parameter

\begin{equation}
\alpha_j = s_j + 1.
\end{equation}

The group allocation is defined using the deterministic Dirichlet mean

\begin{equation}
\bar{p}_j
=
\frac{\alpha_j}{\sum_k \alpha_k}
=
\frac{s_j+1}{\sum_k(s_k+1)}.
\end{equation}

This parameterization maps group-wise vulnerability scores to a valid probability vector on the simplex.

\subsection{Entropy-Regularized Allocation View}

Let

\begin{equation}
u_j = \log \alpha_j
\end{equation}

and let

\begin{equation}
\Delta
=
\left\{
p\mid
p_j\geq0,\,
\sum_jp_j=1
\right\}
\end{equation}

denote the probability simplex. The allocation can equivalently be expressed as the solution of

\begin{equation}
p^\star
=
\arg\max_{p\in\Delta}
\left[
\sum_j p_j u_j
+
H(p)
\right],
\end{equation}

where

\begin{equation}
H(p)
=
-\sum_j p_j\log p_j
\end{equation}

is Shannon entropy. The unique closed-form solution is

\begin{equation}
p_j^\star
=
\frac{\exp(u_j)}
{\sum_k\exp(u_k)}
=
\frac{\alpha_j}
{\sum_k\alpha_k}
=
\bar{p}_j.
\end{equation}

Thus, the deterministic allocation used in CogniDir can be interpreted as an entropy-regularized group-allocation rule with a unique solution.

\subsection{Allocation Properties}

Because $\alpha_j=s_j+1>0$, every group receives positive probability,

\begin{equation}
\bar{p}_j>0.
\end{equation}

The probabilities satisfy

\begin{equation}
\sum_j\bar{p}_j=1.
\end{equation}

Let $S=\sum_k\alpha_k$. The response of a group's allocation to its own vulnerability score is

\begin{equation}
\frac{\partial \bar{p}_j}{\partial s_j}
=
\frac{S-\alpha_j}{S^2}
>0,
\end{equation}

showing that increasing the vulnerability score of group $j$, while holding the other group scores fixed, increases its training allocation.

These properties guarantee simplex feasibility, positive allocation, and monotonic response to group vulnerability. They do not imply a globally optimal training-data mixture for the downstream learning problem.

\subsection{Relation to Dirichlet Modeling}

The Dirichlet distribution is used as a convenient parameterization of the group probability simplex. CogniDir does not define a Bayesian prior over model parameters, perform prior-posterior updating, or sample a posterior distribution. Training uses the deterministic Dirichlet mean rather than random draws from $\mathrm{Dir}(\alpha)$.

Moreover, the implemented allocation does not introduce an independently tuned Dirichlet concentration hyperparameter. The group parameters are determined directly from the vulnerability scores at each training round.

Given a total comment budget $M$, the expected group allocation is converted into integer sample counts using

\begin{equation}
M_j
=
\left\lfloor
\bar{p}_jM
\right\rfloor,
\end{equation}

with any remaining samples assigned according to the largest fractional remainders.

\section{Baselines and Evaluation Protocols}
\label{app:baselines}

We compare against two categories of baselines: (1) LLM-only zero-shot detectors and (2) deep-learning comment-aware fake-news detection models. All baselines receive the news article and its associated comment set as input and output a binary prediction (\texttt{Fake} or \texttt{Real}).

\subsection{LLM-Only Baselines}

We evaluate four instruction-following large language models in a zero-shot setting without task-specific finetuning on the target datasets:

\begin{itemize}
  \item \textbf{Gemma-2-2B} \citep{team2024gemma}: A lightweight open-weight language model designed for efficient language understanding and generation.
  \item \textbf{Mistral-7B} \citep{jiang2023mistral}: An efficient 7B-parameter language model using grouped-query attention and sliding-window attention.
  \item \textbf{Llama-3-8B} \citep{grattafiori2024llama}: An instruction-following language model with strong general language and reasoning capabilities.
  \item \textbf{Qwen2.5-32B} \citep{yang2024qwen25}: A multilingual instruction-following language model with strong reasoning and structured-generation capabilities.
\end{itemize}

\subsection{Inference Reproducibility}

The main experiments use a temperature of $0.2$ for all LLM-only baselines. To examine sensitivity to decoding randomness, we additionally evaluate all four models using greedy decoding ($T=0$) while keeping the prompts, input samples, and evaluation protocol unchanged.

As shown in Tab.~\ref{tab:llm_temperature}, the overall conclusions remain stable across the two decoding settings. The mean Macro-F1 differences are $0.004$, $0.005$, and $0.033$ on Weibo16, Weibo20, and RumourEval-19, respectively, with an overall mean difference of $0.014$.

\begin{table}[t]
\centering
\small
\setlength{\tabcolsep}{3.5pt}
\caption{Macro-F1 of LLM-only baselines under temperature $0.2$ and greedy decoding ($T=0$).}
\label{tab:llm_temperature}
\resizebox{\columnwidth}{!}{
\begin{tabular}{llcccc}
\toprule
Dataset & Model & $T$ & O & A & R \\
\midrule

\multirow{8}{*}{Weibo16}
& \multirow{2}{*}{Gemma-2-2B}  & 0.2 & 0.322 & 0.291 & 0.274 \\
&                              & 0   & 0.514 & 0.292 & 0.309 \\
\addlinespace[1.5pt]
& \multirow{2}{*}{Mistral-7B}  & 0.2 & 0.686 & 0.334 & 0.495 \\
&                              & 0   & 0.668 & 0.372 & 0.401 \\
\addlinespace[1.5pt]
& \multirow{2}{*}{Llama-3-8B}  & 0.2 & 0.598 & 0.324 & 0.408 \\
&                              & 0   & 0.483 & 0.356 & 0.384 \\
\addlinespace[1.5pt]
& \multirow{2}{*}{Qwen2.5-32B} & 0.2 & 0.831 & 0.282 & 0.716 \\
&                              & 0   & 0.805 & 0.495 & 0.529 \\

\midrule

\multirow{8}{*}{Weibo20}
& \multirow{2}{*}{Gemma-2-2B}  & 0.2 & 0.345 & 0.292 & 0.326 \\
&                              & 0   & 0.523 & 0.296 & 0.321 \\
\addlinespace[1.5pt]
& \multirow{2}{*}{Mistral-7B}  & 0.2 & 0.654 & 0.315 & 0.375 \\
&                              & 0   & 0.583 & 0.387 & 0.420 \\
\addlinespace[1.5pt]
& \multirow{2}{*}{Llama-3-8B}  & 0.2 & 0.582 & 0.363 & 0.408 \\
&                              & 0   & 0.423 & 0.337 & 0.357 \\
\addlinespace[1.5pt]
& \multirow{2}{*}{Qwen2.5-32B} & 0.2 & 0.823 & 0.367 & 0.541 \\
&                              & 0   & 0.797 & 0.485 & 0.515 \\

\midrule

\multirow{8}{*}{RumourEval-19}
& \multirow{2}{*}{Gemma-2-2B}  & 0.2 & 0.294 & 0.238 & 0.275 \\
&                              & 0   & 0.441 & 0.238 & 0.258 \\
\addlinespace[1.5pt]
& \multirow{2}{*}{Mistral-7B}  & 0.2 & 0.468 & 0.184 & 0.252 \\
&                              & 0   & 0.549 & 0.310 & 0.339 \\
\addlinespace[1.5pt]
& \multirow{2}{*}{Llama-3-8B}  & 0.2 & 0.377 & 0.219 & 0.276 \\
&                              & 0   & 0.469 & 0.199 & 0.226 \\
\addlinespace[1.5pt]
& \multirow{2}{*}{Qwen2.5-32B} & 0.2 & 0.553 & 0.343 & 0.412 \\
&                              & 0   & 0.523 & 0.352 & 0.377 \\

\bottomrule
\end{tabular}
}
\end{table}

\subsection{Deep-Learning Baselines}

We include five representative comment-aware deep-learning baselines:

\begin{itemize}
  \item \textbf{dEFEND} \citep{shu2019defend}: An explainable model using sentence-comment co-attention to jointly model news and comment evidence.
  \item \textbf{Dual-CAN} \citep{yang2023entity}: An entity-aware dual co-attention network that models news content, user replies, and complementary evidence.
  \item \textbf{GenFEND} \citep{nan2024let}: A generative feedback-enhanced detector that leverages LLM-generated comments to enrich detection evidence.
  \item \textbf{L-Defense} \citep{wang2024explainable}: An LLM-based explainable detector that extracts competing evidence for authenticity prediction.
  \item \textbf{ARG} \citep{hu2024bad}: A representative comment-aware fake-news detection baseline included in our evaluation.
\end{itemize}

\subsection{Evaluation Protocol}

Throughout the evaluation, O denotes performance on the original comment set. A denotes performance on attacked inputs using a model without robust training. R denotes performance after robust training on the same attacked test inputs used for A. Therefore, A and R are evaluated under matched adversarial inputs.

\subsection{Metrics}
\label{app:metrics}

\begin{itemize}

    \item \textbf{Accuracy (Acc):} Overall correctness on the test set.

    \begin{equation}
    \mathrm{Acc}
    =
    \frac{1}{N}
    \sum_{i=1}^{N}
    \mathbb{I}
    [\hat{y}_i=y_i].
    \end{equation}

    \item \textbf{Macro-F1:} Unweighted average of the class-wise F1 scores.

    \begin{equation}
    \mathrm{Macro}\text{-}F1
    =
    \frac{1}{2}
    \left(
    F1_{\mathrm{Real}}
    +
    F1_{\mathrm{Fake}}
    \right).
    \end{equation}

    \item \textbf{Attack Success Rate (ASR):} Fraction of originally correctly classified instances that become misclassified after malicious comments are introduced.

    \begin{equation}
    \mathrm{ASR}
    =
    \frac{
    \sum_{i=1}^{N}
    \mathbb{I}
    \left[
    \hat{y}_i^{O}=y_i
    \wedge
    \hat{y}_i^{A}\neq y_i
    \right]
    }{
    \sum_{i=1}^{N}
    \mathbb{I}
    \left[
    \hat{y}_i^{O}=y_i
    \right]
    }.
    \end{equation}

\end{itemize}

\section{Statistical Analysis}
\label{app:multi-run}

\subsection{Three-Run Stability}

We conduct two additional independent runs for CogniDir and GenFEND under matched experimental settings. Together with the original run, Tab.~\ref{tab:multi_run} reports the mean and sample standard deviation over three runs.

\begin{table}[t]
\centering
\small
\setlength{\tabcolsep}{3.5pt}
\caption{Mean $\pm$ sample standard deviation over three runs.}
\label{tab:multi_run}
\resizebox{\columnwidth}{!}{
\begin{tabular}{llccc}
\toprule
Dataset & Model & O & A & R \\
\midrule

\multirow{2}{*}{Weibo16}
& GenFEND & $0.921\pm0.027$ & $0.577\pm0.025$ & $0.862\pm0.019$ \\
& CogniDir & $\mathbf{0.954\pm0.010}$ & $\mathbf{0.789\pm0.020}$ & $\mathbf{0.943\pm0.011}$ \\
\midrule

\multirow{2}{*}{Weibo20}
& GenFEND & $0.889\pm0.020$ & $0.642\pm0.013$ & $0.761\pm0.021$ \\
& CogniDir & $\mathbf{0.934\pm0.012}$ & $\mathbf{0.729\pm0.020}$ & $\mathbf{0.947\pm0.012}$ \\
\midrule

\multirow{2}{*}{RumourEval-19}
& GenFEND & $0.662\pm0.017$ & $0.457\pm0.024$ & $0.603\pm0.020$ \\
& CogniDir & $\mathbf{0.823\pm0.015}$ & $\mathbf{0.549\pm0.019}$ & $\mathbf{0.810\pm0.015}$ \\

\bottomrule
\end{tabular}
}
\end{table}

CogniDir achieves a higher three-run mean in all nine dataset-setting combinations. The relatively small standard deviations also indicate that the main performance trends are stable across runs.

\subsection{Paired Significance Tests}

We further conduct two-sided paired $t$-tests over the three matched runs.

\begin{table}[t]
\centering
\small
\setlength{\tabcolsep}{6pt}
\caption{Two-sided paired $t$-tests between CogniDir and GenFEND.}
\label{tab:paired_tests}
\begin{tabular}{llcc}
\toprule
Dataset & Setting & $\Delta$ & $p$-value \\
\midrule
\multirow{3}{*}{Weibo16}
& O & +0.033 & 0.089 \\
& A & +0.212 & $<0.001$ \\
& R & +0.080 & 0.042 \\
\midrule

\multirow{3}{*}{Weibo20}
& O & +0.045 & 0.011 \\
& A & +0.087 & 0.003 \\
& R & +0.185 & 0.010 \\
\midrule

\multirow{3}{*}{RumourEval-19}
& O & +0.161 & 0.012 \\
& A & +0.092 & 0.063 \\
& R & +0.207 & 0.009 \\
\bottomrule
\end{tabular}
\end{table}

All nine mean differences are positive. Seven of the nine comparisons are statistically significant at the uncorrected $p<0.05$ level. The improvements on Weibo16-O and RumourEval-19-A remain positive but do not reach this significance threshold.



\end{document}